# Distributed and Adaptive Algorithms for Vehicle Routing in a Stochastic and Dynamic Environment

Marco Pavone, Emilio Frazzoli, Francesco Bullo

## Abstract

In this paper we present distributed and adaptive algorithms for motion coordination of a group of $m$ autonomous vehicles. The vehicles operate in a convex environment with bounded velocity and must service demands whose time of arrival, location and on-site service are stochastic; the objective is to minimize the expected system time (wait plus service) of the demands. The general problem is known as the $m$-vehicle Dynamic Traveling Repairman Problem ($m$-DTRP). The best previously known control algorithms rely on centralized a-priori task assignment and are not robust against changes in the environment, e.g. changes in load conditions; therefore, they are of limited applicability in scenarios involving ad-hoc networks of autonomous vehicles operating in a time-varying environment.

First, we present a new class of policies for the 1-DTRP problem that: (i) are provably optimal both in light- and heavy-load condition, and (ii) are adaptive, in particular, they are robust against changes in load conditions. Second, we show that partitioning policies, whereby the environment is partitioned among the vehicles and each vehicle follows a certain set of rules in its own region, are optimal in heavy-load conditions. Finally, by combining the new class of algorithms for the 1-DTRP with suitable partitioning policies, we design distributed algorithms for the $m$-DTRP problem that (i) are spatially distributed, scalable to large networks, and adaptive to network changes, (ii) are within a constant-factor of optimal in heavy-load conditions and stabilize the system in any load condition. Simulation results are presented and discussed.

## I. INTRODUCTION

Advances in computer technology and wireless communications, coupled with new perceived critical needs of our society, motivate the rapidly increasing interest in the design and deployment of large networks of autonomous mobile vehicles capable of interacting *directly* with the environment [1]. Potential applications are numerous and include surveillance and exploration missions, where vehicles are required to visit points of interest, and transportation and distribution tasks, where vehicles are required to provide service or goods to people. In this context, one of the major scientific challenges is the design of possibly distributed coordination policies to efficiently assign vehicles to *demands* (or *targets*) that are spatially dispersed over the environment.

In the recent past, considerable efforts have been devoted to the problem of how to assign, cooperatively, vehicles to demands [2]–[7]. In these papers, the underlying mathematical model is *static* and fits within the framework of

---

Marco Pavone and Emilio Frazzoli are with the Laboratory for Information and Decision Systems, Department of Aeronautics and Astronautics, Massachusetts Institute of Technology, Cambridge, `{pavone, frazzoli}@mit.edu`.

Francesco Bullo is with the Center for Control Engineering and Computation, University of California at Santa Barbara, `bullo@engineering.ucsb.edu`

the Vehicle Routing Problem (see [8] for a thoroughly introduction to the VRP), whereby: (i) a team of $m$ vehicles is required to service a set of $n$ demands in a 2-dimensional space; (ii) associated with a demand there is an on-site service time; (iii) the goal is to compute a set of routes that optimizes the cost of servicing (according to some quality of service metric) the demands. The VRP is *static* in the sense that vehicles' routes are computed assuming that no new demands arrive.

Despite its richness, inherent elegance and frequent occurrence in practical problems, the VRP is often a *static approximation* to problems that are in reality *dynamic*, and therefore it fails to be an appropriate model in many applications of interest. In fact, demands often arrive *sequentially* in time, and planning algorithms should actually provide *policies* (in contrast to pre-planned routes) that prescribe how the routes should evolve as a function of those inputs that evolve in real-time. From a methodological point of view, dynamic demands (i.e., demands that vary over time) add *queueing phenomena* to the combinatorial nature of the VRP. As a motivating example, consider a surveillance mission where a team of Unmanned Aerial Vehicles (UAVs) must ensure continued coverage of a certain area; as events occur, i.e., as new targets are detected by on-board sensors or other assets (e.g., intelligence, high-altitude or orbiting platforms, etc.), UAVs must proceed to the location of the new event and provide close-range information about the target. Then, the UAV mission control is a *continuous* process of detecting targets, planning routes and sending UAVs. In such a dynamic setting, stability of the overall system is an additional issue to be addressed.

In the present work, we wish to address the assignment problem in the more general framework where new demands for service are generated *over time* by a *stochastic* process. In [9], the authors present a setting where demands arrive sequentially over time and routes are generated in real-time, however the optimization problem is over a finite horizon and the analysis only addresses a number of vehicle $m \leq 2$. Arguably, the most general model for vehicle routing problems that have both a *dynamic* and a *stochastic* component is the $m$-vehicle Dynamic Traveling Repairman Problem ($m$-DTRP) [10]–[13]. In the $m$-DTRP, $m$ agents operating in a convex environment and traveling with bounded velocity must service demands whose time of arrival, location and on-site service are stochastic; the objective is to find a policy to service demands over an infinite horizon that minimizes the expected system time (wait plus service) of the demands. The best previously known control policies for the $m$-DTRP rely on *centralized* task assignment and are *not* robust against changes in the environment, in particular changes in load conditions; therefore, they are of limited applicability in scenarios involving ad-hoc networks of autonomous vehicles operating in a time-varying environment.

In this paper we focus on *unbiased* policies for the DTRP, i.e., policies that provide the same quality of service to all demands independently of their locations. The contribution is threefold: First, we present a new class of unbiased policies for the 1-DTRP problem. In particular, we propose the Divide & Conquer (DC) policy, whose performance depends on a design parameter $r$. If $r \to +\infty$, the policy is (i) provably optimal both in light and heavy-load conditions, and (ii) adaptive with respect to changes in the load conditions and in the statistics of the on-site service requirement; if, instead, $r = 1$, the policy is (i) provably optimal in light-load condition and within a factor of 2 from optimality in heavy-load condition, and (ii) adaptive with respect to all problem data, in particular, and perhaps surprisingly, it does not require *any* knowledge about the demand generation process. Moreover, by applying ideas of receding-horizon control to dynamic vehicle routing problems, we introduce the Receding Horizon (RH) policy, that also does not require any knowledge about the demand generation process; we will show that the RH policy is optimal in light-load and stable in any load condition, and we will heuristically argue that its performance is close to optimality in heavy-load condition. Second, we show that partitioning policies, whereby the environment is partitioned among the vehicles and each vehicle follows a certain set of rules in its own region, are optimal in heavy-load conditions. Finally, by combining our new class of algorithms for the 1-DTRP with suitable partitioning policies, we design distributed algorithms for the $m$-DTRP problem that (i) are spatially distributed, scalable to large networks, and adaptive to network changes, (ii) are within a constant-factor of optimal in heavy-load conditions and stabilize the system in *any* load condition. Here, by network changes we mean changes in the number of vehicles, in the environment boundaries, in the arrival rate of demands, and in the characterization of the on-site service requirement.

The paper is structured as follows. In Section II we provide some background on probability theory, on the Euclidean Traveling Salesman Problem, and on the continuous multi-median problem. In Section III we formulate the problem we wish to address, and we review existing results. In Sections IV and V we present and analyze, respectively, the Divide & Conquer policy and the Receding Horizon policy. In Section VI, building upon the two previous policies for the 1-DTRP, we discuss partitioning policies and we design distributed algorithms for the $m$-DTRP. In Section VII we present results from numerical experiments, and finally, in Section VIII we draw some conclusions and discuss some directions for future work.

## II. Preliminaries

In this section, we briefly describe some known concepts from probability and locational optimization, on which we will rely extensively later in the paper.

### A. Jensen's Inequality

Suppose $h(\cdot)$ is a *convex* function and $X$ a random variable, then

$$\mathbb{E}\left[h(X)\right] \geqslant h(\mathbb{E}\left[X\right]),$$

provided both expectations exist.

### B. Equitable Partitions

Let $\mathcal{Q} \subset \mathbb{R}^d$ be a bounded, convex set. An *r-partition* of $\mathcal{Q}$ (where $r \in \mathbb{N}$, $r \geq 1$) is a collection of $r$ closed subsets $\{\mathcal{Q}_k\}_k$ with disjoint interiors whose union is $\mathcal{Q}$. Given a measurable function $f : \mathcal{Q} \to \mathbb{R}_{>0}$ with $\int_{\mathcal{Q}} f(x)\,dx = 1$, an $r$-partition $\{\mathcal{Q}_k\}_k$ is *equitable* with respect to $f$ if $\int_{\mathcal{Q}_k} f(x)\,dx = 1/r$ for all $k \in \{1, \dots, r\}$. Similarly, given two measurable functions $f_j : \mathcal{Q} \to \mathbb{R}_{>0}$, $j \in \{1, 2\}$, with $\int_{\mathcal{Q}} f_j(x)\,dx = 1$, an $r$-partition $\{\mathcal{Q}_k\}_k$ is *simultaneously equitable* with respect to $f_1$ and $f_2$ if $\int_{\mathcal{Q}_k} f_j(x)\,dx = 1/r$ for all $k \in \{1, \dots, r\}$ and $j \in \{1, 2\}$. Theorem 12 in [14] and Corollary 3 in [15] show that, given two measurable functions $f_j : \mathcal{Q} \to \mathbb{R}_{>0}, j \in \{1, 2\}$, with $\int_{\mathcal{Q}} f_j(x) \, dx = 1$, there *always* exists an $r$-partition that is simultaneously equitable with respect to $f_1$ and $f_2$ and where the subsets $\mathcal{Q}_k$ are convex.

## C. Asymptotic and Worst-Case Properties of the Traveling Salesman Problem in the Euclidean Plane

The Euclidean Traveling Salesman Problem (TSP) is formulated as follows: given a set $D$ of $n$ points in $\mathbb{R}^d$, find the minimum-length tour (i.e., cycle that visits all nodes exactly once) of $D$; the length of a tour is the sum of all Euclidean distances on the tour. Let $\mathrm{TSP}(D)$ denote the minimum length of a tour through all the points in $D$; by convention, $\mathrm{TSP}(\emptyset) = 0$. Assume that the locations of the $n$ points are random variables independently and identically distributed in a compact set $\mathcal{Q}$; in [16] it is shown that there exists a constant $\beta_{\mathrm{TSP},d}$ such that, almost surely,

$$\lim_{n \to +\infty} \frac{\mathrm{TSP}(D)}{n^{1-1/d}} = \beta_{\mathrm{TSP},d} \int_{\mathcal{Q}} \bar{f}(q)^{1-1/d} \, dq, \qquad (1)$$

where $\bar{f}$ is the density of the absolutely continuous part of the distribution of the points. For the case $d = 2$, the constant $\beta_{\mathrm{TSP},2}$ has been estimated numerically as $\beta_{\mathrm{TSP},2} \simeq 0.7120 \pm 0.0002$, [17].

Notice that the bound (1) holds for all compact sets: the shape of the set only affects the convergence rate to the limit. According to [18], if $\mathcal{Q}$ is a "fairly compact and fairly convex" set in the plane, then Eq. (1) provides an adequate estimate of the optimal TSP tour length for values of $n$ as low as 15.

Finally, for a certain environment $\mathcal{Q}$, the following *(deterministic)* bound holds on the length of the TSP tour, uniformly on $n$ [16]:

$$\mathrm{TSP}(D) \leq \beta_{\mathcal{Q},d} \, |\mathcal{Q}|^{1/d} \, n^{(d-1)/d}. \qquad (2)$$

where $|\mathcal{Q}|$ is the Lebesgue measure of $\mathcal{Q}$. The price of determinism is that the constant $\beta_{\mathcal{Q},d}$ depends on the shape of the environment and is generally larger than $\beta_{\mathrm{TSP},d}$.

## D. Tools for solving TSPs

The TSP is known to be NP-complete, which suggests that there is no general algorithm capable of finding the optimum tour in an amount of time polynomial in the size of the input. Even though the exact optimal solution of a large TSP can be very hard to compute, several exact and heuristic algorithms and software tools are available for the numerical solution of Euclidean TSPs.

The most advanced TSP solver to date is arguably `concorde` [19]. Heuristic polynomial-time algorithms are available for constant-factor approximations of TSP solutions, among which we mention Christofides' [20]. On a more theoretical side, Arora proved the existence of polynomial-time approximation schemes, providing a $(1 + \varepsilon)$ constant-factor approximation for any $\varepsilon > 0$ [21].

A modified version of the Lin-Kernighan heuristic [22] is implemented in `linkern`; this powerful solver yields approximations in the order of 5% of the optimal tour cost very quickly for many instances. For example, in our numerical experiments on a 2.4 GHz Pentium machine, approximations of random TSPs with 1,000 points typically required about two seconds of CPU time.[1]

In the following, we will present algorithms that require on-line solutions of large TSPs. Practical implementations of the algorithms will rely on heuristics, such as Lin-Kernighan's or Christofides'. If a constant-factor approximation algorithm is used, the effect on the asymptotic performance guarantees of our algorithms can be simply modeled as a scaling of the constant $\beta_{\mathrm{TSP},d}$.

### E. The Continuous Multi-Median Problem

Given a set $\mathcal{Q} \subset \mathbb{R}^d$ and a vector $P = (p_1, \ldots, p_m)$ of $m$ distinct points in $\mathcal{Q}$, the expected distance between a random point $q$, generated according to a probability density function $f$, and the closest point in $P$ is given by

$$H_m(P, \mathcal{Q}) := \mathbb{E}\left[\min_{k \in \{1, \ldots, m\}} \|p_k - q\|\right] = \sum_{i=1}^m \int_{\mathcal{V}_k(P, \mathcal{Q})} \|p_k - q\| f(q) \, dq,$$

where $\mathcal{V}(P, \mathcal{Q}) = (\mathcal{V}_1(P, \mathcal{Q}), \ldots, \mathcal{V}_m(P, \mathcal{Q}))$ is the Voronoi partition of the set $\mathcal{Q}$ generated by the points $P$. In other words, $q \in \mathcal{V}_k(P, \mathcal{Q})$ if $\|q - p_k\| \leq \|q - p_j\|$, for all $j \in \{1, \ldots, m\}$. The set $\mathcal{V}_k$ is referred to as the Voronoi cell of the generator $p_k$. The function $H_m$ is known in the locational optimization literature as the continuous Weber function or the continuous multi-median function; see [23], [24] and references therein.

The $m$-median of the set $\mathcal{Q}$, with respect to the measure induced by $f$, is the global minimizer

$$P_m^*(\mathcal{Q}) = \operatorname*{argmin}_{P \in \mathcal{Q}^m} H_m(P, \mathcal{Q}).$$

We let $H_m^*(\mathcal{Q}) = H_m(P_m^*(\mathcal{Q}), \mathcal{Q})$ be the global minimum of $H_m$. It is straightforward to show that the map $P \mapsto H_1(P, \mathcal{Q})$ is differentiable and strictly convex on $\mathcal{Q}$. Therefore, it is a simple computational task to compute $P_1^*(\mathcal{Q})$. It is convenient to refer to $P_1^*(\mathcal{Q})$ as the median of $\mathcal{Q}$. On the other hand, the map $P \mapsto H_m(P, \mathcal{Q})$ is differentiable (whenever $(p_1, \ldots, p_m)$ are distinct) but not convex, thus making the solution of the continuous $m$-median problem hard in the general case. It is known [23], [25] that the discrete version of the $m$-median problem is NP-hard for $d \geq 2$. The set of critical points of $H_m$ contains all configurations $(p_1, \ldots, p_m)$ with the property that each point $p_k$ is the generator of the Voronoi cell $\mathcal{V}_k(P, \mathcal{Q})$ as well as the median of $\mathcal{V}_k(P, \mathcal{Q})$ (we refer to such Voronoi diagrams as *median* Voronoi diagrams).

## III. Problem Formulation and Existing Results

In this section we formulate the problem we wish to address and we review existing results.

---

[1] Both `concorde` and `linkern` are written in ANSI C and are freely available for academic research use at `http://www.math.princeton.edu/tsp/concorde.html`.

*A. Problem Formulation*

The basic version of the problem we wish to study in this paper is known as the Dynamic Traveling Repairman Problem (DTRP), which was introduced by Psaraftis in [10] and mainly studied by Bertsimas and van Ryzin in [11]–[13]. In this section, we define the DTRP problem and we review existing results.

Let the workspace $\mathcal{Q} \subset \mathbb{R}^d$ be a bounded, convex set, and let $\|\cdot\|$ denote the Euclidean norm in $\mathbb{R}^d$. We define the diameter of $\mathcal{Q}$ as: $\mathrm{diam}(\mathcal{Q}) \doteq \sup\{\|p - q\| \,|\, p, q \in \mathcal{Q}\}$. For simplicity, we will only consider the planar case, i.e., $d = 2$, with the understanding that extensions to higher dimensions are possible. Consider $m$ holonomic vehicles, modeled as point masses, and let

$$p(t) = (p^1(t), \ldots, p^m(t)) \in \mathcal{Q}^m$$

describe the locations of the vehicles at time $t$. The vehicles are free to move, traveling at a constant velocity $v$, within the environment $\mathcal{Q}$. The vehicles are identical, and have unlimited fuel and demand servicing capacity.

Demands are generated according to a homogeneous (i.e., time-invariant) spatio-temporal Poisson process, with time intensity $\lambda$, and spatial density $f : \mathcal{Q} \to \mathbb{R}_{>0}$. In other words, demands arrive to $\mathcal{Q}$ according to a Poisson process with intensity $\lambda$, and their locations $\{X_i;\ i \geqslant 1\}$ are i.i.d. (i.e., *independent* and *identically distributed*) and distributed according to a density $f(\cdot)$ whose support is $\mathcal{Q}$. A demand's location becomes known (is realized) at the demand's arrival epoch; thus, at time $t$ we know with *certainty* the locations of demands that arrived prior to time $t$, but future demand locations form an i.i.d sequence. The notation $f(x)$ is short for $f(x_1, x_2)$ ($x = [x_1, x_2]$). The density $f(\cdot)$ satisfies:

$$\mathbb{P}\left[X_i \in S\right] = \int_S f(x)\,dx \quad \forall S \subseteq \mathcal{Q}, \quad \text{and} \quad \int_\mathcal{Q} f(x)\,dx = 1.$$

We assume that the number of initial outstanding demands is a random variable with finite first and second moments.

At each demand location, vehicles spend some time $s$ in on-site service that is i.i.d. and generally distributed with finite first and second moments denoted by $\bar{s}$ and $\overline{s^2}$. A realized demand is removed from the system after one of the vehicles has completed its on-site service. We define the *load factor* $\varrho \doteq \lambda \bar{s}/m$.

The system time of demand $j$, denoted by $T_j$, is defined as the elapsed time between the arrival of demand $j$ and the time one of the vehicles completes its service. The waiting time of demand $j$, $W_j$, is defined by $W_j = T_j - s_j$. The steady-state system time is defined by $\bar{T} \doteq \limsup_{j \to \infty} \mathbb{E}\left[T_j\right]$. A policy for routing the vehicles is said to be *stable* if the expected number of demands in the system is bounded uniformly at all times. A necessary condition for the existence of a stable policy is that $\varrho < 1$; we shall assume $\varrho < 1$ throughout the paper. When we refer to *light-load* condition, we consider the case $\varrho \to 0^+$, in the sense that $\lambda \to 0^+$; when we refer to *heavy-load* condition, we consider the case $\varrho \to 1^-$, in the sense that $\lambda \to [1/\bar{s}]^-$.

Let $\mathcal{P}$ be the set of all causal, stable and stationary routing policies. Letting $\bar{T}_\pi$, denote the system time of a particular policy $\pi \in \mathcal{P}$, the DTRP is then defined as the problem of finding a policy $\pi^*$ (if one exists) such that

$$\bar{T}_{\pi^*} = \inf_{\pi \in \mathcal{P}} \bar{T}_\pi.$$

We let $\bar{T}^*$ denote the infimum in the right hand side above.

We finally need two definitions; in these definitions, $X$ is the location of a randomly chosen demand and $W$ is its waiting time.

*Definition 3.1:* A policy $\pi$ is called *spatially unbiased* if for every pair of sets $\mathcal{S}_1, \mathcal{S}_2 \subseteq \mathcal{Q}$

$$\mathbb{E}\left[W \,|\, X \in \mathcal{S}_1\right] = \mathbb{E}\left[W \,|\, X \in \mathcal{S}_2\right];$$

a policy $\pi$ is instead called *spatially biased* if there exists sets $\mathcal{S}_1, \mathcal{S}_2 \subseteq \mathcal{Q}$ such that

$$\mathbb{E}\left[W \,|\, X \in \mathcal{S}_1\right] > \mathbb{E}\left[W \,|\, X \in \mathcal{S}_2\right].$$

In this paper we focus on unbiased policies; indeed, unbiased service seems to be a natural constraint in most applications of interest, e.g., surveillance by a team of UAVs. We are interested in *distributed*, *scalable*, *adaptive* control policies with *provable performance guarantees*, that rely only on local exchange of information between neighboring vehicles, and do not require explicit knowledge of the global structure of the network.

### B. Existing Results for the $m$-DTRP

As in many queueing problems, the analysis of the $m$-DTRP problem for all values of $\varrho \in [0, 1)$ is difficult. In [11]–[13], [26], lower bounds for the optimal steady-state system time are derived for the light-load case (i.e., $\varrho \to 0^+$), and for the heavy-load case (i.e., $\varrho \to 1^-$). Subsequently, policies are designed for these two limiting regimes, and their performance is compared to the lower bounds.

*1) Lower Bounds:* For the light-load case, a tight lower bound on the system time is derived in [12]. In the light-load case, the lower bound on the $m$-DTRP system time is strongly related to the solution of the $m$-median problem:

$$\bar{T}^* \geq \frac{1}{v} H_m^*(\mathcal{Q}) + \bar{s}, \qquad \text{as} \qquad \varrho \to 0^+. \tag{3}$$

This bound is tight: there exist policies whose system times, in the limit $\varrho \to 0^+$, attain this bound; we present such asymptotically optimal policies for the light-load case below.

The following two lower bounds hold in the heavy-load case [13], [26]. Within the class of spatially *unbiased* policies the optimal system time is lower bounded by

$$\bar{T}^* \geq \frac{\beta_{\text{TSP},2}^2}{2} \frac{\lambda \left[\int_{\mathcal{Q}} f^{1/2}(x)dx\right]^2}{m^2 \, v^2 \, (1-\varrho)^2} \qquad \text{as} \qquad \varrho \to 1^-. \tag{4}$$

Within the class of spatially *biased* policies the optimal system time is lower bounded by

$$\bar{T}^* \geq \frac{\beta_{\text{TSP},2}^2}{2} \frac{\lambda \left[\int_{\mathcal{Q}} f^{2/3}(x)dx\right]^3}{m^2 \, v^2 \, (1-\varrho)^2} \qquad \text{as} \qquad \varrho \to 1^-. \tag{5}$$

Both bounds (4) and (5) are *tight*: there exist policies whose system times, in the limit $\varrho \to 1^-$, attain these bounds; therefore the inequalities in (4) and (5) should indeed be replaced by equalities. We present asymptotically optimal policies for the heavy-load case below. It is shown in [13] that the lower bound in Eq. (5) is always lower than or equal to the lower bound in Eq. (4) for all densities $f$.

 

*Remark 3.2:* In equations (4) and (5), the right-hand side approaches $+\infty$ as $\varrho \to 1^-$. Thus, one should more formally write the inequalities with $\bar{T}^*(1-\varrho)^2$ on the left-hand side, so that the right-hand side is finite. However, this makes the presentation less readable, and thus, in the remainder of the paper we adhere to the less formal but more transparent style of equations (4) and (5). Furthermore, this is the style generally used in the literature on the DTRP (see, e.g., [13] Section 4 and [27] Proposition 2).

*Remark 3.3:* It is possible to show (see [13], Proposition 1) that a *uniform* spatial density function is the *worst possible* and that any deviation from uniformity in the demand distribution will strictly lower the optimal mean system time in both the unbiased and biased case. In addition, allowing biased service will result in a strict reduction of the optimal mean system time for any non-uniform density $f$. Also, when the density is uniform there is nothing to be gained by not providing unbiased service.

*2) Optimal Policies for the Light Load Case:* For the light-load case, to the best of our knowledge, there is only one policy that is known to be optimal; this policy is as follows.

> **The $m$ Stochastic Queue Median (SQM) Policy** [12] — Locate one vehicle at each of the $m$ median locations for the environment $\mathcal{Q}$. When demands arrive, assign them to the nearest median location and its corresponding vehicle. Have each vehicle service its respective demands in First-Come, First-Served (FCFS) order returning to its median after each service is completed.

This policy, although optimal in light load, has several drawbacks: (i) it does not stabilize the system as the load increases, (ii) it relies on the solution of the continuous $m$-median problem that is hard for $m > 1$, (iii) it relies on *a priori* computation and, thus, is not adaptive to changes in the environment, and, finally, (iv) it requires a *centralized* assignment to the median locations.

*3) Optimal Policies for the Heavy Load Case:* For the heavy-load case, to the best of our knowledge, there is only one unbiased policy that has been proven to achieve the lower bound in (4); this policy is as follows.

> **The Unbiased TSP (UTSP) Policy** [13] — Let $r$ be a fixed positive, large integer. From a central point in the interior of $\mathcal{Q}$, subdivide the service region into $r$ wedges $\mathcal{Q}_1, \mathcal{Q}_2, \cdots, \mathcal{Q}_r$ such that $\int_{\mathcal{Q}_i} f(x)dx = 1/r$, $i = 1, 2, \cdots, r$. Within each subregion, form sets of size $n/r$ ($n$ is a design parameter). As sets are formed, deposit them in a queue and service them FCFS with the first available vehicle by forming a TSP on the set and following it in an arbitrary direction. Optimize over $n$.

It is possible to show that

$$\bar{T}_{\text{UTSP}} \leq \left(1 + \frac{m}{r}\right) \frac{\beta_{\text{TSP},2}^2}{2} \frac{\lambda \left[\int_{\mathcal{Q}} f^{1/2}(x)dx\right]^2}{m^2 \, v^2 \, (1-\varrho)^2} \qquad \text{as} \qquad \varrho \to 1^-.$$

Thus, letting $r \to \infty$, the lower bound in (4) is achieved.

The same paper [13] presents an optimal biased policy. This policy, called Biased TSP (BTSP) Policy, relies on an even finer partition of the environment and requires $f$ to be piecewise constant.

Although both policies are optimal within their respective classes, they have the following drawbacks: (i) they do not perform well in light load, (ii) the parameter $n$ is selected a priori and depends on the data of the problem,

 

(iii) they require a centralized data structure (the demands' queue is shared by the vehicles), (iv) the partition of the environment is computed a priori and clearly depends on $f$.

### C. Toward Distributed, Scalable, and Adaptive Control Policies for the $m$-DTRP

A candidate distributed, scalable, and adaptive control policy for the $m$-DTRP is the simple Nearest Neighbor (NN) policy: at each service completion epoch, each vehicle chooses to visit next the closest unserviced demand, if any, otherwise it stops at the current position. Because of the dependencies among the travel inter-demand distances, the analysis of the NN policy is difficult and no rigorous results have been obtained so far [11]; in particular, its stability properties are not known. Simulation experiments show that the NN policy performs like a biased policy and is not optimal *neither* in the light-load case *nor* in the heavy-load case. In particular, when $f$ is uniform, in light load the steady-state system time under the NN policy is 40% larger than the steady-state system time under the (optimal) SQM policy, while in heavy load the steady-state system time under the NN policy is 60% larger than the steady-state system time under the (optimal) Biased TSP policy [11], [13]. Therefore, the NN policy *lacks* provable performance guarantees (in particular about stability), is *biased* and does *not* seem to achieve optimal performance neither in light load nor in heavy load.

The key idea that we will pursue in this paper is that of *partitioning policies*.

*Definition 3.4:* Given a policy $\pi$ for the 1-DTRP and $m$ vehicles, a $\pi$-partitioning policy is a family of multi-vehicle policies such that

1) the environment $\mathcal{Q}$ is partitioned into $m$ openly disjoint subregions $\mathcal{Q}_k$, $k \in \{1, \ldots, m\}$, whose union is $\mathcal{Q}$,

2) one vehicle is assigned to each subregion (thus, there is a bijective correspondence between vehicles and subregions),

3) each vehicle executes the single-vehicle policy $\pi$ to service demands that fall within its own subregion.

Note that Definition 3.4 does not specify how the environment is actually partitioned, therefore Definition 3.4 describes a *family* of policies (one for each partitioning strategy) for the $m$-DTRP. The SQM policy, which is optimal *in light load*, is indeed a partitioning policy whereby $\mathcal{Q}$ is partitioned according to a median Voronoi diagram that "minimizes" $H_m$, and each vehicle executes inside its own Voronoi cell the single-vehicle policy "service FCFS and return to the median after each service completion". Moreover, in Section VI, we will prove that *in heavy load*, given a single-vehicle, unbiased optimal policy $\pi^*$, there exists an unbiased $\pi^*$-partitioning policy that is optimal.

The above discussion leads to the following strategy: First, for the 1-DTRP, we design unbiased, scalable, and adaptive control policies with provable performance guarantees. Then, by using recently developed *distributed* algorithms for environment partitioning, we extend these single-vehicle policies to the multi-vehicle case.

## IV. THE SINGLE-VEHICLE DIVIDE & CONQUER POLICY

We describe the Divide & Conquer (DC) policy in Algorithm 1, where $D$ is the set of outstanding demands and $p$ is the current position of the vehicle. An $r$-partition that is simultaneously equitable with respect to $f(x)$

Check modulo r

---

**Algorithm 1: Divide & Conquer (DC) Policy**

---

**Assumes**: An $r$-partition $\{\mathcal{Q}_k\}_{k=1}^{r}$ of $\mathcal{Q}$ that is simultaneously equitable with respect to $f(x)$ and

$$f^{1/2}(x)/\int_{\mathcal{Q}} f^{1/2}(x)\ dx.$$

**1 if** $D = \emptyset$ **then**

**2**      Let $\tilde{P}_1^*$ be the point minimizing the sum of distances to demands serviced in the past (the initial condition for $\tilde{P}_1^*$ is a random point in $\mathcal{Q}$); if $p \neq \tilde{P}_1^*$, move to $\tilde{P}_1^*$, otherwise stop.

**3 else**

**4**      $k \leftarrow k_o$. // $k_0$ is uniformly randomly chosen in the set $\{1, \ldots, r\}$.

**5**      **repeat**

**6**          Move to the median of subregion $\mathcal{Q}_k$.

**7**          Compute the TSP tour through all demands in subregion $\mathcal{Q}_k$. Service all demands by following the TSP tour, starting at a random point on the TSP tour.

**8**          $k \leftarrow k + 1$ modulo $r$.

**9**      **until** $D = \emptyset$

**10** Repeat.

---

and $f^{1/2}(x)/\int_{\mathcal{Q}} f^{1/2}(x)\ dx$ is indeed guaranteed to exist (see discussion on equitable partitions in Section II). The number of subregions $r$ is a design parameter whose choice will be discussed after the analysis of the policy. One should note that $\tilde{P}_1^*$ is indeed the empirical median. Next, we characterize the DC policy. Notice that the DC policy is an *unbiased* policy.

### A. Analysis of the DC Policy in Light Load

We first study the light-load case (i.e., $\varrho \to 0^+$).

*Theorem 4.1 (Performance of DC policy in light load):* As $\varrho \to 0^+$, for every $r$, the DC policy is asymptotically optimal, that is,

$$\bar{T}_{DC}(r) \to \bar{T}^*, \qquad \text{as } \varrho \to 0^+.$$

*Proof:* Using arguments similar to those in [27], consider generic initial conditions in $\mathcal{Q}$ for the vehicle's position and for the positions of the outstanding demand (denote the initial number of demands with $n_0$). Under the DC policy, an upper bound to the time $C_0$ needed to service *all* of the initial demands and move to the empirical median $\tilde{P}_1^*$ is: $C_0 \leq (n_0 + 1)\,\text{diam}(\mathcal{Q})/v + \sum_{j=1}^{n_0} s_j$. Thus, we have $\mathbb{E}\left[C_0\right] \leq \mathbb{E}\left[n_0\right](\text{diam}(\mathcal{Q})/v + \bar{s}) + \text{diam}(\mathcal{Q})/v$.

Given time interval $C_0 = t$, $t \in \mathbb{R}_{\geq 0}$, the probability that no *new* demand arrives during such time interval is $\mathbb{P}\left(N(C_0) = 0 | C_0 = t\right) = \exp(-\lambda t)$, where $N(t)$ is the counting variable associated to a Poisson process with rate

$\lambda$. Hence, by applying the law of total probability and Jensen's inequality for convex functions

$$\mathbb{P}\left[N(C_0) = 0\right] = \int_0^\infty \mathbb{P}\left(N(C_0) = 0 | C_0 = t\right) f_{C_0}(t) dt = \int_0^\infty \exp(-\lambda t) f_{C_0}(t) dt \geq \exp\left(-\lambda \int_0^\infty t f_{C_0}(t) dt\right)$$

$$= \exp\left(-\lambda \mathbb{E}\left[C_0\right]\right) \geq \exp\left(-\lambda \mathbb{E}\left[n_0\right](\operatorname{diam}(\mathcal{Q})/v + \bar{s}) - \lambda \operatorname{diam}(\mathcal{Q})/v\right), \quad (6)$$

where $f_{C_0}$ is the density function of random variable $C_0$. Since, by assumption, $\mathbb{E}\left[n_0\right] < \infty$ and $\varrho \to 0^+$ (i.e., $\lambda \to 0^+$), we obtain $\mathbb{P}\left[N(C_0) = 0\right] \to 1^-$ as $\varrho \to 0^+$. As a consequence, after an initial transient, all demands will be generated with the vehicle at the empirical median $\tilde{P}_1^*$, and an empty demand queue, with probability 1.

Thus, after an initial transient, the DC policy, in light load, *reduces* to line 2 with the condition $D = \emptyset$ true with probability 1. Then, in light load, the DC policy becomes identical to the light-load routing policy presented by one of the authors in [28]; since such routing policy is optimal, in light load, for the 1-DTRP [28], we obtain the desired result (the proof in [28] basically shows that $\tilde{P}_1^* \to P_1^*$). ∎

### B. Analysis of the DC Policy in Heavy Load

Next, we consider the DC policy in heavy load, i.e., when $\varrho \to 1^-$. The performance of the DC policy in heavy load is characterized by the following theorem.

*Theorem 4.2 (Performance of DC policy in heavy load):* As $\varrho \to 1^-$, the system time for the DC policy satisfies

$$\bar{T}_{\text{DC}}(r) \leq \left(1 + \frac{1}{r}\right) \frac{\beta_{\text{TSP},2}^2}{2} \frac{\lambda\left[\int_Q f^{1/2}(x)\,dx\right]^2}{v^2\,(1 - \varrho)^2}, \quad (7)$$

where $r$ is the number of subregions.

To prove Theorem 4.2 we use techniques similar to those we developed in [29], [30]. We start the analysis with the following Lemma, similar to Lemma 1 in [27], characterizing the number of outstanding demands in heavy load.

*Lemma 4.3 (Number of outstanding demands in heavy load for DC policy):* In heavy load (i.e., $\varrho \to 1^-$), after a transient, the number of demands serviced in a single tour of the vehicle in the DC policy is very large with high probability (i.e., the number of demands tends to $+\infty$ with probability that tends to 1, as $\varrho$ approaches $1^-$).

*Proof:* Consider the case where the vehicle moves with infinite velocity (i.e., $v \to +\infty$); then the system is reduced to the usual M/G/1 queue. The infinite-velocity system has fewer demands waiting in queue. A known result on M/G/1 queues [31] states that, after transients, the total number of demands, as $\varrho \to 1^-$, is very large with high probability. Thus, the number of outstanding demands in each subregion is also very large with high probability. In particular, after transients, the number of demands is very large with high probability at the time instants when the vehicle starts a new TSP tour. ∎

Lemma 4.3 has two implications. First, since the number of demands is very large at the time instants when the vehicle starts a new TSP tour, we can apply Eq. (1) to estimate the length of a TSP tour. Second, since $D \neq \emptyset$ with high probability, the DC policy, in heavy load, *reduces* to lines 5-9 of Algorithm 1 with the condition $D = \emptyset$ *always* false.

We refer to the time instant $t_i$, $i \geq 0$, in which the vehicle starts a new TSP tour in subregion 1 as the epoch $i$ of the policy; we refer to the time interval between epoch $i$ and epoch $i+1$ as the $i$th iteration. Let $n_i^k$, $k \in \{1, \ldots, r\}$, be the number of outstanding demands serviced in subregion $k$ during iteration $i$. Finally, let $C_i^k$ be the time interval between the time instant the vehicle starts to service demands in subregion $k$ during iteration $i$ and the time instant the vehicle starts to service demands in the same subregion $k$ during next iteration $i+1$. Demands arrive in subregion $\mathcal{Q}_k$ according to a Poisson process with rate $\hat{\lambda} \doteq \lambda/r$, where we use the fact that the partition $\{\mathcal{Q}_k\}_{k=1}^r$ is equitable with respect to $f$; then, we have $\mathbb{E}\left[n_{i+1}^k\right] = \hat{\lambda}\,\mathbb{E}\left[C_i^k\right]$. The time interval $C_i^k$ is the sum of three components:

1) the time for the $r$ travels from the end of one TSP tour to the start of next one;

2) the time to perform $r$ TSP tours;

3) the time to provide on-site service while performing the $r$ TSP tours.

Assume that $i$ is large enough (say, $i \geq \bar{\imath}$) so that, according to Lemma 4.3, the number of outstanding demands is large. The first component is trivially upper bounded by $r\operatorname{diam}(\mathcal{Q})/v$. Given that a demand falls in subregion $\mathcal{Q}_k$, the conditional density for its location (whose support is $\mathcal{Q}_k$) is $f(x)/\int_{\mathcal{Q}_k} f(x)\,dx$; then, the expected length of a TSP tour through $n_i^k$ demands (with a slight abuse of notation, we call such length $TSP(n_i^k)$) can be upper bounded as

$$\mathbb{E}\left[TSP(n_i^k)\right] \leq \beta_{\text{TSP},2}\left(\int_{\mathcal{Q}_k} \sqrt{\frac{f(x)}{\int_{\mathcal{Q}_k} f(x)\,dx}}\,dx\right)\sqrt{\mathbb{E}\left[n_i^k\right]} = \beta_{\text{TSP},2}\,\sqrt{r}\,\frac{\int_{\mathcal{Q}} f^{1/2}(x)\,dx}{r}\,\sqrt{\mathbb{E}\left[n_i^k\right]} \qquad (8)$$

where we use Eq. (1), we apply Jensen's inequality for concave functions in the form $\mathbb{E}\left[\sqrt{X}\right] \leq \sqrt{\mathbb{E}\left[X\right]}$, and we exploit the fact that, by definition of the DC policy, the partition $\{\mathcal{Q}_k\}_{k=1}^r$ is simultaneously equitable with respect to $f(x)$ and $f^{1/2}(x)/\int_{\mathcal{Q}} f^{1/2}(x)\,dx$. Define $\beta \doteq \beta_{\text{TSP},2}\frac{\int_{\mathcal{Q}} f^{1/2}(x)\,dx}{\sqrt{r}}$; equation (8) becomes $\mathbb{E}\left[TSP(n_i^k)\right] \leq \beta\sqrt{\mathbb{E}\left[n_i^k\right]}$. Finally, since on-site service times are independent of the number of outstanding demands, the expected on-site service requirement for $n_i^k$ demands is $\bar{s}\,\mathbb{E}\left[n_i^k\right]$.

Then, we obtain the following recurrence relation (where we define $\bar{n}_i^k \doteq \mathbb{E}\left[n_i^k\right]$):

$$\bar{n}_{i+1}^k = \hat{\lambda}\mathbb{E}\left[C_i^k\right] \leq \hat{\lambda}\left(r\operatorname{diam}(\mathcal{Q})/v + \frac{\beta}{v}\sum_{j=k}^r \sqrt{\bar{n}_i^j} + \frac{\beta}{v}\sum_{j=1}^{k-1}\sqrt{\bar{n}_{i+1}^j} + \bar{s}\sum_{j=k}^r \bar{n}_i^j + \bar{s}\sum_{j=1}^{k-1}\bar{n}_{i+1}^j\right), \quad k \in \{1,\dots,r\}. \qquad (9)$$

The $r$ inequalities above describe a system of recurrence relations that allows to find an upper bound on $\limsup_{i \to +\infty} \bar{n}_i^k$. The following Lemma bounds the values to which the limits $\limsup_{i \to +\infty} \bar{n}_i^k$ converge.

*Lemma 4.4 (Steady state number of demands for DC policy):* In heavy load, for every set of initial conditions $\{\bar{n}_1^k\}_{k \in \{1,\dots,r\}}$, the trajectories $i \mapsto \bar{n}_i^k$, $k \in \{1,\dots,r\}$, resulting from Eq. (9), satisfy

$$\bar{n}^k \doteq \limsup_{i \to +\infty} \bar{n}_i^k \leq \beta_{\text{TSP},2}^2\,\frac{\lambda^2\left[\int_{\mathcal{Q}} f^{1/2}(x)\,dx\right]^2}{r\,v^2\,(1-\varrho)^2}$$

*Proof:* By Lemma 4.3, $n_i^j$ tends to $+\infty$ with probability that tends to 1, as $\varrho$ approaches $1^-$; then, after a transient, the term $r\operatorname{diam}(\mathcal{Q})/v$ is negligible compared to the other terms in the right hand side of Eq. (9), and therefore it can be ignored (its inclusion in the proof is conceptually straightforward, but makes the analysis more cumbersome).

 

Next we define two auxiliary systems, System-Y and System-Z. We define System-Y (with state $y(i) \in \mathbb{R}^r$) as

$$y_k(i+1) = \hat{\lambda}\left( \sum_{j=k}^{r}\left( \bar{s}\, y_j(i) + \frac{\beta}{v}\sqrt{y_j(i)} \right) + \sum_{j=1}^{k-1}\left( \bar{s}\, y_j(i+1) + \frac{\beta}{v}\sqrt{y_j(i+1)} \right) \right), \qquad k \in \{1,\ldots,r\}. \tag{10}$$

System-Y is obtained by replacing the inequality in Eq. (9) with an equality. Pick, now, any $\varepsilon > 0$. From Young's inequality

$$\sqrt{a} \leq \frac{1}{4\varepsilon} + \varepsilon a, \quad \text{for all } a \in \mathbb{R}_{\geq 0}.$$

By applying Young inequality in Eq. (10) we obtain

$$y_k(i+1) \leq \hat{\lambda}\left( \bar{s} + \varepsilon\frac{\beta}{v} \right)\left( \sum_{j=k}^{r} y_j(i) + \sum_{j=1}^{k-1} y_j(i+1) \right) + r\,\frac{\hat{\lambda}\beta}{4v\varepsilon}, \qquad k \in \{1,\ldots,r\}. \tag{11}$$

Next, define System-Z (with state $z(i) \in \mathbb{R}^r$) as

$$z_k(i+1) = \hat{\lambda}\left( \bar{s} + \varepsilon\frac{\beta}{v} \right)\left( \sum_{j=k}^{r} z_j(i) + \sum_{j=1}^{k-1} z_j(i+1) \right) + r\,\frac{\hat{\lambda}\beta}{4v\varepsilon}, \qquad k \in \{1,\ldots,r\}, \tag{12}$$

The proof now proceeds as follows. First, we show that if $\bar{n}_{\bar{i}}^k \leq y_k(\bar{i}) \leq z_k(\bar{i})$ for all $k$, then

$$\bar{n}_i^k \leq y_k(i) \leq z_k(i), \quad \text{for all } k \text{ and for all } i \geq \bar{i}. \tag{13}$$

Second, we show that the trajectories of System-Z are bounded; this fact, together with Eq. (13), implies that also trajectories of variables $\bar{n}_i^k$ and $y_k(i)$ are bounded. Third, and last, we will compute $\limsup_{i \to +\infty} y_k(i)$; this quantity, together with Eq. (13), will yield the desired result.

Let us consider the first issue. Given $i$, we prove that if $\bar{n}_i^k \leq y_k(i) \leq z_k(i)$ for all $k \in \{1,\ldots,r\}$, then $\bar{n}_{i+1}^k \leq y_k(i+1) \leq z_k(i+1)$. Indeed, it is immediate to show that, under the assumptions, it holds: $\bar{n}_{i+1}^1 \leq y_1(i+1) \leq z_1(i+1)$; moreover, it is easy to show that, under the assumptions, and if $\bar{n}_{i+1}^j \leq y_j(i+1) \leq z_j(i+1)$ for all $j \leq k < r$, then $\bar{n}_{i+1}^{k+1} \leq y_{k+1}(i+1) \leq z_{k+1}(i+1)$. The claim follows by induction on $k$. Equation (13) is a consequence of the statement we have just proven.

We now turn our attention to the second issue, namely boundedness of trajectories for System-Z (described in Eq. (12)). Define $\delta \doteq \hat{\lambda}(\bar{s} + \varepsilon\,\beta/v)$. System-Z is a discrete-time linear system and can be rewritten (by adding and subtracting $z_{k-1}(i)$ in the second term between parentheses in Eq. (12), when $k > 1$) as

$$z_k(i+1) = \begin{cases} \delta \sum_{j=1}^{r} z_j(i) + r\,\frac{\hat{\lambda}\beta}{4v\varepsilon} & \text{for } k=1, \\ \left(1+\delta\right) z_{k-1}(i+1) - \delta\, z_{k-1}(i) & \text{for } k \in \{2,\ldots,r\}. \end{cases}$$

By using the above equation and by simple induction arguments, it is possible to show that System-Z can be written in canonical form as

$$z_k(i+1) = (1+\delta)^{k-1}\left( \delta \sum_{j=1}^{r} z_j(i) + r\,\frac{\hat{\lambda}\beta}{4v\varepsilon} \right) - \sum_{j=1}^{k-1} \delta(1+\delta)^{k-j-1} z_j(i).$$

Hence, System-Z can be written in compact form as

$$z(i+1) = \Big( A + B(\varepsilon) \Big) z(i) + c(\varepsilon),$$

where $A \in \mathbb{R}^{r \times r}$, $B(\varepsilon) \in \mathbb{R}^{r \times r}$ and each element of $B(\varepsilon)$, say $b(\varepsilon)_{kj}$, has the property $\lim_{\varepsilon \to 0+} b(\varepsilon)_{kj} = 0$ for all $k, j \in \{1, \dots, r\}$, and $c(\varepsilon) \in \mathbb{R}^{r \times 1}$ (notice that $c(\varepsilon)$ is well defined for every $\varepsilon > 0$). It is easy to see that matrix $A$ is

$$a_{kj} = \begin{cases} \hat{\varrho}\Big[(1 + \hat{\varrho})^{k-1} - (1 + \hat{\varrho})^{k-j-1}\Big] & \text{for } 1 \le j \le k-1, \\ \hat{\varrho}(1 + \hat{\varrho})^{k-1} & \text{otherwise,} \end{cases}$$

where $\hat{\varrho} \doteq \hat{\lambda}\bar{s}$. Notice that, since $\hat{\varrho} > 0$, we have $|a_{kj}| = a_{kj}$ for all $k$ and $j$. Then, for each $k \in \{1, \dots, r\}$, we have

$$\sum_{j=1}^{r} |a_{kj}| = \sum_{j=1}^{r} a_{kj} = r\,\hat{\varrho}\,(1 + \hat{\varrho})^{k-1} - \hat{\varrho}\sum_{j=1}^{k-1}(1 + \hat{\varrho})^{k-j-1} = (1 + \hat{\varrho})^{k-1}\underbrace{(r\,\hat{\varrho} - 1)}_{<0} + 1 < 1,$$

since $\hat{\varrho} = \lambda\bar{s}/r$ and $\varrho = \lambda\bar{s} < 1$ by assumption. Therefore, the $L^{\infty}$-induced norm of matrix A satisfies: $\|A\|_{\infty} = \max_k \sum_{j=1}^{r} |a_{kj}| < 1$. Since any induced norm $\|\cdot\|$ satisfies the inequality $\rho(A) \le \|A\|$, where $\rho(A)$ is the spectral radius of $A$, we conclude that $A \in \mathbb{R}^{r \times r}$ has eigenvalues strictly inside the unit disk (i.e., $A$ is a stable matrix). Since the eigenvalues of a matrix depend continuously on the matrix entries, there exists a sufficiently small $\varepsilon > 0$ such that the matrix $A + \varepsilon B$ has eigenvalues strictly inside the unit disk. Accordingly, each solution $i \mapsto z(i) \in \mathbb{R}^{r}_{\ge 0}$ of System-Z converges exponentially fast to the unique equilibrium point

$$z^* = \Big(I_r - A - B(\varepsilon)\Big)^{-1} c(\varepsilon). \tag{14}$$

Combining Eq. (13) with the previous statement, we see that the solutions $i \mapsto \bar{n}(i)$ and $i \mapsto y(i)$ are bounded (where $\bar{n}(i) \doteq (\bar{n}_i^1, \dots, \bar{n}_i^r)$). Thus

$$\limsup_{i \to +\infty} \bar{n}(i) \le \limsup_{i \to +\infty} y(i) < +\infty. \tag{15}$$

Finally, we turn our attention to the third issue, namely the computation of $y \doteq \limsup_{i \to +\infty} y(i)$. Taking the $\limsup$ of the left- and right-hand sides of Eq. (10), and noting that

$$\limsup_{i \to +\infty} \sqrt{y_j(i)} = \sqrt{\limsup_{i \to +\infty} y_j(i)} \quad \text{for } j \in \{1, \dots, r\},$$

since $\sqrt{\cdot}$ is continuous and strictly monotone increasing on $\mathbb{R}_{>0}$, we obtain that

$$y_k = \hat{\lambda} \sum_{j=1}^{r} \Big(\bar{s}\,y_j + \frac{\beta}{v}\sqrt{y_j}\Big); \tag{16}$$

therefore we have $y_k = y_j$ for all $k, j \in \{1, \dots, r\}$. Substituting in Eq. (16) and rearranging we obtain

$$y_k = \frac{\lambda^2 \beta^2}{v^2\,(1 - \varrho)^2}.$$

Noting that from Eq. (15), $\limsup_{i \to +\infty} \bar{n}_i^k \le y_k$, we obtain the desired result. ∎

We are now in a position to prove theorem 4.2

*Proof of Theorem 4.2:* Define $\bar{C}^k \doteq \limsup_{i \to \infty} \mathbb{E}\left[C_i^k\right]$; then we have, by using the upper bound on $\mathbb{E}\left[C_i^k\right]$ in Eq. (9),

$$\bar{C}^k \doteq \limsup_{i \to \infty} \mathbb{E}\left[C_i^k\right] \le \sum_{j=1}^{r} \frac{\beta}{v}\sqrt{\bar{n}^j} + \bar{s}\sum_{j=1}^{r} \bar{n}^j \le \frac{r\,\lambda\,\beta^2}{v^2\,(1 - \varrho)} + \frac{\varrho\,r\,\lambda\beta^2}{v^2\,(1 - \varrho)^2}.$$

Hence, in the limit $\varrho \to 1^-$, we have

$$\bar{C}^k \le \frac{r\,\lambda\,\beta^2}{v^2\,(1-\varrho)^2} = \frac{\lambda\beta_{\mathrm{TSP},2}^2 \left[\int_Q f^{1/2}(x)\,dx\right]^2}{v^2\,(1-\varrho)^2}.$$

Consider a random demand that arrives in subregion $k$. Its expected steady-state system time, $\bar{T}^k$, will be upper bounded, as $\varrho \to 1^-$, by

$$\bar{T}^k \le \frac{1}{2}\,\bar{C}^k + \frac{1}{2}\,\bar{s}\,\bar{n}^k \le \frac{1}{2}\,\bar{C}^k + \frac{1}{2}\,\frac{\lambda\beta^2}{v^2\,(1-\varrho)^2},$$

where we used the fact that, as $\varrho \to 1^-$, the travel time along a TSP tour is negligible compared to the on-site service time requirement. Unconditioning, we obtain the claim

$$\bar{T} \le \left(1 + \frac{1}{r}\right) \frac{\lambda\beta_{\mathrm{TSP},2}^2 \left[\int_Q f^{1/2}(x)\,dx\right]^2}{2v^2\,(1-\varrho)^2}$$

∎

### C. Discussion

The DC policy is optimal in light load (Theorem 4.1); moreover, if we let $r \to \infty$, the DC policy achieves the heavy-load lower bound (4) for unbiased policies (Theorem 4.2). Therefore the DC policy is *both* optimal in light load *and* arbitrarily close to optimality in heavy load, and stabilizes the system in every load condition (since it stabilizes the system in the limit $\varrho \to 1^-$). Notice that with $r = 10$ the DC policy is already guaranteed to be within 10% of the optimal (for unbiased policies) performance in heavy load.

The DC policy does not rely on any *parameter* that is a function of the problem data: $r$ is simply a design factor whose choice is independent of $\lambda, \bar{s}, v$ and $f$. However, if $r > 1$, the computation of a simultaneously equitable $r$-partition of $Q$ does depend on $f$. Therefore, if $r \to +\infty$, the policy is (i) provably optimal both in light and heavy load, and (ii) adaptive with respect to arrival rate $\lambda$, expected on-site service $\bar{s}$, and vehicle's velocity $v$. If, instead, $r = 1$, the policy is (i) provably optimal in light load and within a factor of 2 of optimal performance in heavy load, and (ii) adaptive with respect to arrival rate $\lambda$, expected on-site service $\bar{s}$, vehicle's velocity $v$, and spatial density $f$; in other words, when $r = 1$, the DC policy adapts to *all* problem data.

An approximate algorithm to compute, with *arbitrarily small* error, an $r$-partition that is simultaneously equitable with respect to two measures can be found in [14] (specifically, see discussion on page 621); in the particular case when the density $f$ is uniform, the problem of finding an $r$-partition that is simultaneously equitable with respect to $f(x)$ and $f^{1/2}(x)/\int_Q f^{1/2}(x)\,dx$ becomes trivial. Given a simultaneously equitable $r$-partition of $Q$, the subregions can be indexed using the following procedure: (i) a TSP tour through the medians of the subregions is computed, (ii) subregions are indexed according to the order induced by this TSP tour. In practice, the DC policy would be implemented by allowing the vehicle to skip subregions where no demand is present and by directly moving from one TSP to the next one, i.e., by skipping the median of subregion $Q_k$ in line 6 (the "shortcuts" were not included in the presentation of the DC policy to make its analysis easier).

Note that if we assume that $f$ is known (as it must be the case for the implementation of the DC policy with $r > 1$), then instead of using the empirical mean $\tilde{P}_1^*$ one should directly use the exact median $P_1^*$ (which is the solution of a simple convex problem).

Finally, the DC policy with $r = 1$ is similar to the generation policy presented in [27]; in particular, the generation policy has the same performance guarantees of the DC policy with $r = 1$. However, the generation policy is analyzed only for the case of *uniform* spatial density $f$, and its implementation *does* require the knowledge of $f$ (while, as discussed before, the DC policy with $r = 1$ adapts to all problem data, including $f$). The part of Algorithm 1 in lines 5-9 is similar to the PART-TSP policy presented in [26]; however, the PART-TSP policy is only *informally* analyzed by assuming steady-state conditions, in particular no proof of stability and convergence is provided.

In the next section we present and analyze another single-vehicle policy, the Receding Horizon (RH) policy, that applies ideas of receding-horizon control to dynamic vehicle routing problems.

## V. The Single-Vehicle Receding Horizon Policy

Next, we describe the Receding Horizon (RH) policy. In what follows, $D$ is the set of outstanding demands and $p$ is the current position of the vehicle. Moreover, given a tour $T$ of $D$, a fragment of $T$ is a connected subgraph of $T$.

---

**Algorithm 2: Receding Horizon (RH) Policy**

---

**input**: Scalar $\eta \in (0, 1)$

1  **if** $D = \emptyset$ **then**

2       Let $\tilde{P}_1^*$ be the point minimizing the sum of distances to demands serviced in the past; if $p \neq \tilde{P}_1^*$, move to $\tilde{P}_1^*$, otherwise stop.

3  **else**

4       **repeat**

5           Compute the TSP tour through all demands in $D$.

6           Uniformly randomly choose, independently from the past, a fragment of length $\eta\, TSP(D)$ of such TSP tour.

7           Service the selected fragment starting from the endpoint closest to the current position (if the entire tour is selected, start from the closest demand).

8       **until** $D = \emptyset$

9  Repeat.

---

In other words, if $D \neq \emptyset$, the vehicle selects a random $\eta$-fragment of the TSP tour through the demands in $D$ (i.e., a fragment of length $\eta\, TSP(D)$ of such tour); note that the horizon is not fixed, but it is adjusted according to the cost of servicing the outstanding demands. In general, the performance of the system will depend on the choice of the parameter $\eta$, which will be discussed after the analysis of the policy. Notice that the RH policy is an *unbiased* policy.

### A. Stability and Performance of the RH Policy

The RH policy is attractive since it can be implemented without any knowledge of the underlying demand generation process (in particular, without any knowledge of $f$). On the other hand, an important caveat of receding horizon control is that closed-loop stability is not guaranteed in general. In this section, we study the stability of the RH policy and we discuss its performance.

*1) Stability of the RH Policy:* In general, the stability properties of receding horizon control deteriorate as the horizon becomes shorter; remarkably, the RH policy is stable in every load for every $0 < \eta < 1$.

*Theorem 5.1:* As $\varrho \to 1^-$, the system time for the RH policy satisfies

$$\bar{T}_{\text{RH}}(\eta) \leq \frac{\lambda \beta_{\mathcal{Q},2}^2 |\mathcal{Q}|}{v^2 (1 - \varrho)^2}, \qquad \text{for all } \eta \in (0, 1),$$

where $\beta_{\mathcal{Q},2}$ is the constant appearing in Eq. (2).

*Proof:* By following the same arguments as in Lemma 4.3, it is easy to show that under the RH policy, after a transient, the number of demands at the instants when a new TSP tour is computed is very large with high probability. Then, in heavy load, $D \neq \emptyset$ with high probability, and the RH policy *reduces* to lines 4-8 with the condition $D = \emptyset$ *always* false.

We refer to the time instant $t_i$, $i \geq 0$, in which the vehicle computes a new TSP tour as the epoch $i$ of the policy; we refer to the time interval between epoch $i$ and epoch $i + 1$ as the $i$th iteration and we refer to its length as the cost $C_i$. Let $n_i$ be the number of outstanding demands at epoch $i$; with a slight abuse of notation, we denote by $TSP(n_i)$ the length of the TSP tour through the outstanding demands at epoch $i$.

As in the proof of Lemma 4.4, the heavy-load assumption allows us to neglect the time to travel from the end of one TSP tour to the start of next one. Since an $\eta$-fragment is randomly chosen, the expected number of demands left unserviced during iteration $i$ is $(1 - \eta) \, \mathbb{E}\,[n_i]$; then,

$$\mathbb{E}\,[n_{i+1}] = \underbrace{\lambda \, \mathbb{E}\,[C_i]}_{\text{newly arrived demands}} + \underbrace{(1 - \eta) \, \mathbb{E}\,[n_i]}_{\text{demands left unserviced during iteration } i}. \qquad \text{Clear, right?}$$

The expected number of demands that receive service during iteration $i$ is $\eta \, \mathbb{E}\,[n_i]$. Then, the expected time-length of iteration $i$ is given by

$$\mathbb{E}\,[C_i] = \bar{s} \, \eta \mathbb{E}\,[n_i] + \frac{\eta}{v} \, \mathbb{E}\,[TSP(n_i)] \leq \bar{s} \, \eta \mathbb{E}\,[n_i] + \frac{\eta}{v} \, \beta_{\mathcal{Q},2} \sqrt{|\mathcal{Q}|} \, \sqrt{\mathbb{E}\,[n_i]}, \tag{17}$$

where we have applied the deterministic bound in Eq. (2) and Jensen's inequality for concave functions. Therefore, we obtain

$$\mathbb{E}\,[n_{i+1}] \leq \varrho \, \eta \, \mathbb{E}\,[n_i] + \lambda \, \frac{\eta}{v} \, \beta_{\mathcal{Q},2} \sqrt{|\mathcal{Q}|} \, \sqrt{\mathbb{E}\,[n_i]} + (1 - \eta) \, \mathbb{E}\,[n_i].$$

It is then straightforward to show that

$$\bar{n} \doteq \limsup_{i \to \infty} \, \mathbb{E}\,[n_i] \leq \frac{\lambda^2 \beta_{\mathcal{Q},2}^2 |\mathcal{Q}|}{v^2 (1 - \varrho)^2}, \quad \text{for all} \quad \eta \in (0, 1).$$

From Little's law, $\bar{n} = \lambda \bar{T}_{RH}$, and thus

$$\bar{T}_{RH} \leq \frac{\lambda \beta_{\mathcal{Q},2}^2 |\mathcal{Q}|}{v^2 (1-\varrho)^2} \quad \text{for all} \quad \eta \in (0,1).$$

∎

Since, by Theorem 5.1, the RH policy stabilizes the system in the limit $\varrho \to 1^-$, it also stabilizes the system for *any* load $\varrho \in [0,1)$.

*2) Performance of the RH Policy:* In light load, the RH policy becomes identical to the DC policy, and therefore it is *optimal* (see Theorem 4.1).

For the case $\varrho \to 1^-$, because of the dependencies among demands' locations, we were unable to obtain upper bounds on the system time that are both tight and rigorous. However, we now present an heuristic analysis of the RH policy that provides interesting insights on its behavior and suggests bounds on its heavy-load system time that are well matched by simulation results.

The service of an $\eta$-fragment introduces *dependencies* among the locations of the demands that are left unserviced; therefore, even though the number of demands, as $\varrho \to 1^-$, becomes large with probability 1, the result in Eq. (1) (that requires a set of *independently* distributed demands) formally does *not* hold. However, due to the randomized selection of the $\eta$-fragments, it is quite natural to conjecture that Eq. (1) still provides a good estimate on the lengths of the TSP tours that are computed. The rationale behind this conjecture is the following: When $\eta$ is close to 1, most of the outstanding demands are serviced during one iteration, and the next TSP tour is mostly through newly arrived demands, that are independent by the assumptions on the demand generation process; when, instead, $\eta$ is close to zero, the randomized selection of *short* fragments only introduce "negligible" correlations.

The above observations motivate us to use $\beta_{\mathrm{TSP},2} \int_Q f^{1/2}(x)\, dx$ (as it would be dictated by Eq. (1) ) in Eq. (17), instead of $\beta_{\mathcal{Q},2} \sqrt{|\mathcal{Q}|}$ (as it is dictated by Eq. (2)). Then, by repeating the same steps as in the proof of Theorem 5.1, we would obtain the following result

$$\bar{T}_{\mathrm{RH}}(\eta) \leq \frac{\lambda \beta_{\mathrm{TSP},2}^2 \left[ \int_Q f^{1/2}(x)\, dx \right]^2}{v^2 (1-\varrho)^2}, \qquad \text{for all } \eta \in (0,1). \tag{18}$$

The upper bound in (18) is tighter than the upper bound in Theorem 5.1 (notice that by Jensen's inequality for convex functions $\left[ \int_Q f^{1/2}(x)\, dx \right]^2 \leq \int_Q (f^{1/2})^2(x)\, dx = 1$), but unlike the upper bound in Theorem 5.1, it has not been established formally. Simulation results (see Section VII) indeed confirm the upper bound in Eq. (18).

Finally, note that when $\eta$ is close to zero, the RH policy is conceptually similar to the DC policy with $r \to \infty$, so we also speculate that in heavy load the performance of the RH policy improves as the horizon becomes shorter. Simulation results (see Section VII) indeed confirm this behavior.

## B. Discussion

The RH policy is optimal in light load and stabilizes the system in every load condition (these two statements have been established rigorously). The implementation of the RH policy does *not* require the knowledge of $\lambda, \bar{s}, v$ and $f$. Therefore, the RH policy *adapts* to arrival rate $\lambda$, expected on-site service $\bar{s}$, vehicle's velocity $v$, and spatial density $f$; in other words, the RH policy adapts to *all* problem data.

In Section IV we saw that also the DC policy with $r = 1$ adapts to all problem data. Simulation results (see Section VII) show that in heavy load the RH policy with a short horizon $\eta$ (say, $\eta \approx 0.2$) performs better than the DC policy with $r = 1$ (the light-load behavior is clearly the same). Therefore, to date, the best available policy for the 1-DTRP that is unbiased and adapts to all problem data seems to be the RH policy with $\eta \approx 0.2$.

It is natural to wonder how the RH policy performs if the $\eta$-fragment is not selected randomly, but it is instead selected according to the rule: Find the fragment of length $\eta \, \mathrm{TSP}(D)$ that maximizes the number of reached demands. In other words, the vehicle now looks for a maximum-reward $\eta$-fragment. Clearly, Theorem 5.1 still applies. Simulation results (see [32]) show that this variant of the RH policy appears to behave neither like a pure unbiased policy nor like an optimal biased policy; rather, its performance seems to lie between these two extremes.

## VI. Adaptive and Distributed Policies for the $m$-DTRP

In this section we extend the previous single-vehicle policies to the multi-vehicle case. First, we prove that in heavy load, given a single-vehicle, unbiased optimal policy $\pi^*$, there exists an unbiased $\pi^*$-partitioning policy that is optimal.

This result and the optimality of the SQM policy (which is a partitioning policy that is optimal in light load) lead us to propose, for the multi-vehicle case, *partitioning policies* that use the DC policy or the RH policy as single-vehicle policies. Finally, we discuss how a partition can be computed in a distributed fashion, so that the proposed policies are amenable to distributed implementation.

### A. Optimality of Partitioning Policies in Heavy Load

The following theorem holds.

*Theorem 6.1:* In heavy load, given a single-vehicle, unbiased optimal policy $\pi^*$ and $m$ vehicles, there exists an unbiased $\pi^*$-partitioning policy that is optimal for the $m$-DTRP.

*Proof:* Let $\pi^*$ be an optimal unbiased policy (e.g., the DC policy with $r \to \infty$). Construct an $m$-partition $\{\mathcal{Q}_i\}_{i=1}^m$ of $\mathcal{Q}$ that is simultaneously equitable with respect to $f(x)$ and $f^{1/2}(x)/\int_{\mathcal{Q}} f^{1/2}(x)\,dx$ (such partition is guaranteed to exist as discussed in Section II); assign one vehicle to each subregion. Each vehicle executes the single-vehicle unbiased policy $\pi^*$ to service demands that fall within its own subregion. The probability that a demand falls in subregion $\mathcal{Q}_i$ is equal to $\int_{\mathcal{Q}_i} f(x)\,dx = 1/m$. Notice that the arrival rate $\lambda_i$ in subregion $\mathcal{Q}_i$ is reduced by a factor $\int_{\mathcal{Q}_i} f(x)\,dx$, i.e., $\lambda_i = \lambda \int_{\mathcal{Q}_i} f(x)\,dx = \lambda/m$; therefore, the load factor in subregion $\mathcal{Q}_i$ (where only one vehicle provides service) is $\varrho_i = \lambda_i \bar{s} = \lambda \bar{s}/m = \varrho < 1$, in particular the necessary condition for stability *is satisfied* in every subregion. Finally, given that a demand falls in subregion $\mathcal{Q}_i$, the conditional density for its location (whose support is $\mathcal{Q}_i$) is $f(x)/\int_{\mathcal{Q}_i} f(x)\,dx = m\,f(x)$. Clearly, such multi-vehicle policy is still unbiased.

Then, by the law of total probability we have

$$\bar{T} = \sum_{i=1}^{m} \left( \int_{\mathcal{Q}_i} f(x)\,dx \; \frac{\beta_{\mathrm{TSP},2}^2}{2} \frac{\lambda_i}{v^2\,(1-\varrho_i)^2} \left[ \int_{\mathcal{Q}_i} \sqrt{\frac{f(x)}{\int_{\mathcal{Q}_i} f(x)\,dx}}\,dx \right]^2 \right)$$

$$= \sum_{i=1}^{m} \left( \frac{1}{m} \; \frac{\beta_{\mathrm{TSP},2}^2}{2} \frac{\lambda}{v^2\,(1-\varrho)^2} \left[ \int_{\mathcal{Q}_i} f^{1/2}(x)\,dx \right]^2 \right)$$

$$= \frac{\beta_{\mathrm{TSP},2}^2}{2} \frac{\lambda \left[ \int_{\mathcal{Q}} f^{1/2}(x)\,dx \right]^2}{m^2\,v^2\,(1-\varrho)^2} \qquad \text{as} \qquad \varrho \to 1^-.$$

Thus, the bound in (4) is achieved.

Check equality ∎

*Remark 6.2:* Theorem 6.1 proves Conjecture 1 (made for uniform spatial density $f$) in [12].

*Remark 6.3:* A similar result can be proven for biased policies; in this case the $m$-partition should be simultaneously equitable with respect to $f(x)$ and $f^{2/3}/\int_{\mathcal{Q}} f^{2/3}(x)\,dx$).

check

*Remark 6.4:* Interestingly the proof of Theorem 6.1 uses, for unbiased policies, $m$-partitions that are simultaneously equitable with respect[2] to $f$ and $f^{1/2}$. We next show that a $\pi^*$-partitioning policy that uses a partition equitable with respect to $f^{1/2}$ might be unstable, while a $\pi^*$-partitioning policy that uses a partition equitable with respect to $f$ is in general not optimal but has a constant factor guarantee of $m$.

*Theorem 6.5:* In heavy load, given a single-vehicle, unbiased optimal policy $\pi^*$, a $\pi^*$-partitioning policy that uses an $m$-partition that is equitable with respect to $f^{1/2}$ may be unstable.

*Proof:* The arrival rate in subregion $\mathcal{Q}_i$ is $\lambda_i = \lambda \int_{\mathcal{Q}_i} f(x)\,dx$, and therefore the load factor in subregion $\mathcal{Q}_i$ (where only one vehicle provides service) is $\varrho_i = \lambda \int_{\mathcal{Q}_i} f(x)\,dx\,\bar{s}$. In general, an $m$-partition that is equitable with respect to $f^{1/2}$ is *not* equitable with respect to $f$ (unless the density is uniform); if this is the case, let $\mathcal{Q}_{\bar{i}}$ be the subregion such that $\int_{\mathcal{Q}_{\bar{i}}} f(x)\,dx\,\bar{s} = 1/m + \varepsilon$, for some $\varepsilon > 0$; then, $\varrho_{\bar{i}} = \varrho + \varepsilon\lambda\bar{s}$. Therefore, as $\varrho \to 1^-$, demands in subregion $\mathcal{Q}_{\bar{i}}$ will experience unbounded system times (for values of $\varrho$ strictly less than one). ∎

*Theorem 6.6:* In heavy load, given a single-vehicle, unbiased optimal policy $\pi^*$, a $\pi^*$-partitioning policy that uses an $m$-partition that is equitable with respect to $f$ is within a factor $m$ of the optimal, and in general is not optimal.

*Proof:* Let $\pi^*$ be an optimal unbiased policy. The environment $\mathcal{Q}$ in partitioned into $m$ subregions $\mathcal{Q}_i$, $i = 1, \ldots, m$, such that $\int_{\mathcal{Q}_i} f(x)\,dx = \int_{\mathcal{Q}} f(x)\,dx/m = 1/m$ for all $i$, and one vehicle is assigned to each subregion. Each vehicle executes the single-vehicle policy $\pi^*$ to service demands that fall within its own subregion. Clearly, such multi-vehicle policy is unbiased. As in the proof of Theorem 6.1 we have, by the law of total probability

$$\bar{T} = \sum_{i=1}^{m} \left( \int_{\mathcal{Q}_i} f(x)\,dx \; \frac{\beta_{\mathrm{TSP},2}^2}{2} \frac{\lambda_i}{v^2\,(1-\varrho_i)^2} \left[ \int_{\mathcal{Q}_i} \sqrt{\frac{f(x)}{\int_{\mathcal{Q}_i} f(x)\,dx}}\,dx \right]^2 \right)$$

$$= \sum_{i=1}^{m} \left( \frac{1}{m} \; \frac{\beta_{\mathrm{TSP},2}^2}{2} \frac{\lambda}{v^2\,(1-\varrho)^2} \left[ \int_{\mathcal{Q}_i} f^{1/2}(x)\,dx \right]^2 \right) \qquad \text{as} \qquad \varrho \to 1^-. \tag{19}$$

---

[2]With a slight abuse of notation we are omitting the normalization constant.

Define the vector

$$x = \left( \int_{\mathcal{Q}_1} f^{1/2}(x)\, dx, \int_{\mathcal{Q}_2} f^{1/2}(x)\, dx, \ldots, \int_{\mathcal{Q}_m} f^{1/2}(x)\, dx \right);$$

we will denote by $\|x\|_2$ and $\|x\|_1$ the 2-norm and the 1-norm, respectively, of vector $x$. Notice that $\|x\|_1 = \int_{\mathcal{Q}} f^{1/2}(x)\, dx$. Since for any vector $x \in \mathbb{R}^m$ it holds $\|x\|_1 \leq \sqrt{m}\|x\|_2$, Eq. (19) becomes

$$\bar{T} = \frac{1}{m} \frac{\beta_{\mathrm{TSP},2}^2}{2} \frac{\lambda}{v^2\,(1-\varrho)^2} \|x\|_2^2 \geq \frac{1}{m} \frac{\beta_{\mathrm{TSP},2}^2}{2} \frac{\lambda}{v^2\,(1-\varrho)^2} \frac{\|x\|_1^2}{m}$$

$$= \frac{\beta_{\mathrm{TSP},2}^2}{2} \frac{\lambda \left[ \int_{\mathcal{Q}} f^{1/2}(x)\, dx \right]^2}{m^2\,v^2\,(1-\varrho)^2} = \bar{T}^* \qquad \text{as} \qquad \varrho \to 1^-.$$

In general, an $m$-partition that is *equitable* with respect to $f$ is *not* equitable with respect to $f^{1/2}$; therefore, the inequality $\|x\|_1 \leq \sqrt{m}\|x\|_2$ is, in general, strict and it can happen that $\bar{T} > \bar{T}^*$, i.e., that such partitioning policy is not optimal.

On the other hand, for any vector $x \in \mathbb{R}^m$ it also holds $\|x\|_2 \leq \|x\|_1$, therefore Eq. (19) becomes

$$\bar{T} = \frac{1}{m} \frac{\beta_{\mathrm{TSP},2}^2}{2} \frac{\lambda}{v^2\,(1-\varrho)^2} \|x\|_2^2 \leq \frac{1}{m} \frac{\beta_{\mathrm{TSP},2}^2}{2} \frac{\lambda}{v^2\,(1-\varrho)^2} \|x\|_1^2$$

$$= \frac{\beta_{\mathrm{TSP},2}^2}{2} \frac{\lambda \left[ \int_{\mathcal{Q}} f^{1/2}(x)\, dx \right]^2}{m\,v^2\,(1-\varrho)^2} = \bar{m}\, T^* \qquad \text{as} \qquad \varrho \to 1^-.$$

∎

*Remark 6.7:* A similar result can be proven for biased policies: in this case the 3-norm is the relevant norm to use.

### B. Distributed Policies for the m-DTRP and Discussion

The optimality of the SQM policy and Theorem 6.1 *suggest* the following distributed multi-vehicle version of the DC policy:

1) the vehicles compute in a distributed way an $m$-median of $\mathcal{Q}$ that induces a Voronoi partition that is equitable with respect to $f$ and $f^{1/2}$,

2) the vehicles assign themselves to the subregions (so that there is a bijective correspondence between vehicles and subregions),

3) each vehicle executes the single-vehicle DC policy to service demands that fall within its own subregion, by using the median of the subregion instead of $\tilde{P}_1^*$.

(In an identical way, we can obtain a multi-vehicle version of the RH policy.)

*If*, given $\mathcal{Q}$ and $f$, an $m$-median of $\mathcal{Q}$ that induces a Voronoi partition that is equitable with respect to $f$ and $f^{1/2}$ exists, then the above policy is optimal both in light load and arbitrarily close to optimality in heavy load, and stabilizes the system in every load condition. There are two main issues with the above policy, namely (i) existence of an $m$-median of $\mathcal{Q}$ that induces a Voronoi partition that is equitable with respect to $f$ and $f^{1/2}$, and (ii) how to compute it in a distributed way. In [33], we showed that for some choices of $\mathcal{Q}$ and $f$ a median Voronoi diagram that is equitable with respect to $f$ and $f^{1/2}$ *fails* to exist. On the other hand, in [34], we presented a synchronous distributed partitioning algorithm that, for any possible choice of $\mathcal{Q}$ and $f$, provides a partition that is equitable with respect to $f$ and represents a "good" approximation of a median Voronoi diagram (see [34] for details on the metrics that we use to judge "closeness" to median Voronoi diagrams). Moreover, if an $m$-median of $\mathcal{Q}$ that induces a Voronoi partition that is equitable with respect to $f$ exists, the algorithm will locally converge to it.

Accordingly, we define the multi-vehicle Divide & Conquer policy as follows

---

**Algorithm 3**: **Multi-Vehicle DC ($m$-DC) Policy**

---

**1** The vehicles run the distributed partitioning algorithm in [34].

**2** The vehicles assign themselves to the subregions (this part is indeed a by-product of the distributed algorithm [34]).

**3** Each vehicle executes the single-vehicle DC policy inside its own subregion.

---

According to Theorem 6.6 the $m$-DC policy is within $m$ of optimal in heavy load (since the algorithm in [34] *always* provides a partition that is equitable with respect to $f$), and stabilizes the system in every load condition. In general, the $m$-DC policy is only *suboptimal* in light load; note, however, that the computation of the global minimum of the Weber function $H_m$ (which is non-convex for $m > 1$) is difficult for $m > 1$ (it is NP-hard for the discrete version of the problem); therefore, for $m > 1$, suboptimality has to be expected from *any* practical implementation of the SQM policy. If an $m$-median of $\mathcal{Q}$ that induces a Voronoi partition that is equitable with respect to $f$ exists, the $m$-DC will locally converge to it, thus we say that the $m$-DC policy is "locally" optimal in light load.

Note that, when the density is uniform, a partition that is equitable with respect to $f$ is also equitable with respect to $f^{1/2}$; therefore, when the density is uniform the $m$-DC policy is arbitrarily close to optimality in heavy load (see Theorem 6.1).

The $m$-DC policy *adapts* to arrival rate $\lambda$, expected on-site service $\bar{s}$, and vehicle's velocity $v$; however, it requires the knowledge of $f$. The key feature of the partitioning algorithm in [34] is that it is *spatially-distributed* (see, e.g., [35] for a rigorous definition of spatially-distributed algorithms); therefore, the $m$-DC policy is indeed a distributed control policy.

## VII. Simulation Experiments

In this section we discuss, through the use of simulations, the heavy-load behavior of the DC policy and the RH policy, and their relative performance.

All simulations are performed in Matlab®7.4, with external calls to the program `linkern`. In all simulations we consider a circular environment $\mathcal{Q}$ with area equal to 1, a vehicle's velocity $v = 1$, and an on-site service time uniformly distributed in the interval $[0, 1]$ (thus $\bar{s} = 0.5$). In each simulation run we iterate the policy (either DC or RH) *until a steady-state is reached*; we use the following method to declare that the system is in steady-state: we filter the sequence of iterations' time-lengths by using a moving-average filter with window size 15, and we stop as soon as the linear least-square fit of the previous 300 iterations' time-lengths yields a slope with magnitude less than 0.1. On average, when using as initial condition $n_0 = (\varrho/\bar{s})\,\bar{T}_{\mathrm{DC}}$ (respectively, $n_0 = (\varrho/\bar{s})\,\bar{T}_{\mathrm{RH}}$), a steady-state (according to the previous criterion) is reached after $\sim 1500$ iterations; clearly, larger values of $\varrho$ imply a larger relaxation time. The spatial density $f : \mathcal{Q} \to \mathbb{R}_+$ used in the simulation experiments is illustrated in Fig. 4. The regions $\mathcal{Q}_1$ and $\mathcal{Q}_2$ have areas $\varepsilon$ and $1 - \varepsilon$ respectively. Within each region demands are uniformly distributed. Demands fall in region $\mathcal{Q}_1$ with probability $1 - \delta$ and in region $\mathcal{Q}_2$ with probability $\delta$. Thus, the density is piecewise uniform with:

$$f(x) = \begin{cases} \frac{1-\delta}{\epsilon} & x \in \mathcal{Q}_1, \\[2mm] \frac{\delta}{1-\epsilon} & x \in \mathcal{Q}_2. \end{cases}$$

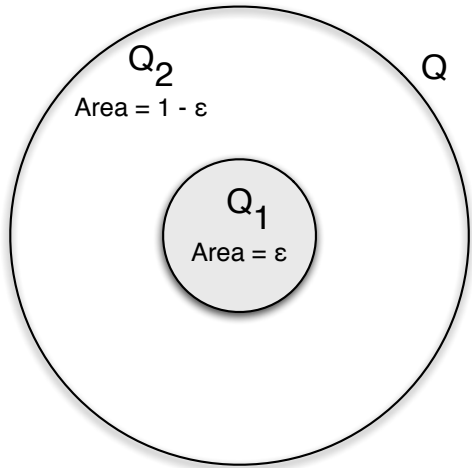

Fig. 1.    Geometry of spatial density $f$.

If $1 - \delta = \varepsilon$, then the density $f$ is uniform; if, instead, $1 - \delta > \varepsilon$, then the density $f$ has a peak in subregion $\mathcal{Q}_1$. In all simulation experiments, we set $\varepsilon = 0.1$.

## A. Heavy-Load Performance of the DC Policy

We first consider a uniform spatial density, i.e., $\delta = 1 - \varepsilon = 0.9$. We consider 5 values of the load factor, namely $\varrho = 0.9, 0.93, 0.95, 0.97, 0.99$. For each value of $\varrho$, we perform 50 runs by executing the DC policy with $r = 1$. In each run we iterate the policy until a steady-state is reached (in the sense described before), and we compute the average system time by considering the last 300 iterations; this value is then averaged over the 50 runs, and is finally divided by $\bar{T}^*$ (whose expression is given in Eq. (4)). Red circles in the left-hand figure of Fig. 2 represent such ratios for the different values of $\varrho$. We then repeat the simulation process by executing the DC policy with $r = 16$. Black squares in the left-hand figure of Fig. 2 represent the ratios between the experimental system time and $\bar{T}^*$ for the different values of $\varrho$. It can be observed that for $r = 1$ the ratios $\bar{T}_{DC}/\bar{T}^*$ decrease as $\varrho$ approaches one and tend to 2, in accordance with the upper bound in Theorem 4.2 (in fact, $1 + 1/r = 2$ in this case). Similarly, for $r = 16$ the ratios $\bar{T}_{DC}/\bar{T}^*$ decrease as $\varrho$ approaches one and tend to $(1 + 1/16) \approx 1.06$, in accordance again with the upper bound in Theorem 4.2. These results confirm the theoretical analysis in Section IV and show that the upper bound (7) is indeed tight for the various values of $r$. Note that for all values of $\varrho$ the experimental system time is *larger* than the upper bound; this is due to one or a combination of the following reasons: First, the upper bound formally holds only in the limit $\varrho \to 1^-$. Second, we are using an approximate solution for the optimal TSP.

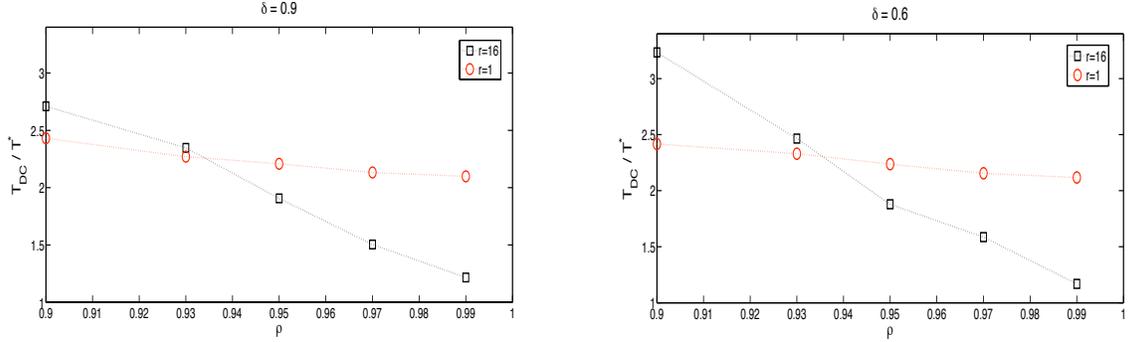

Fig. 2. Left Figure: Ratio between experimental system time under the DC policy and $\bar{T}^*$ in the case of uniform density (i.e., $\delta = 0.9$). Right Figure: Ratio between experimental system time under the DC policy and $\bar{T}^*$ in the case of non-uniform density ($\delta = 0.6$). Red circles correspond to the DC policy with $r = 1$, while black squares correspond to the DC policy with $r = 16$.

The same set of simulations is performed for a non-uniform spatial density; in particular we consider $\delta = 0.6$, i.e., a peak in the small subregion $\mathcal{Q}_1$. Note that in this case a simultaneously equitable partition can be trivially obtained by using radial cuts. Results are shown in the right-hand figure of Fig. 2. Again, red circles represent the ratios between the experimental system time and $\bar{T}^*$ for $r = 1$, while black squares represent the ratios between the experimental system time and $\bar{T}^*$ for $r = 16$. As before, the results are in accordance with the the upper bound in Theorem 4.2.

Finally, from Fig. 2 it can be observed that as $\varrho \to 1^-$ the DC policy with $r = 16$ performs better than the DC policy with $r = 1$, precisely by a factor of 2 in accordance with the upper bound in Theorem 4.2. On the other hand, for moderate values of $\varrho$, say $\varrho \approx 0.9$, the DC policy with $r = 1$ performs better than the DC policy with $r = 16$. This result is expected since for moderate values of $\varrho$ and "large" values of $r$ the number of demands in each subregion is low, and therefore Eq. (1) is no longer applicable (while it is still applicable for $r = 1$).

### B. Heavy-Load Performance of the RH Policy

We first consider a uniform spatial density. We consider 3 values of the load factor, namely $\varrho = 0.95, 0.97, 0.99$, and 7 values for $\eta$, namely $\eta = 0.1, 0.2, 0.3, 0.7, 0.8, 0.9, 1$. Formally, the RH policy is not defined for $\eta = 1$, however, by setting $\eta = 1$ in the definition of the RH policy, we simply obtain the DC policy with $r = 1$. For each pair $(\varrho, \eta)$, we perform, as before, 50 runs by executing the RH policy (with the corresponding $\eta$). In each run we iterate the policy until a steady-state is reached, and we compute the average system time by considering the last 300 iterations; this value is then averaged over the 50 runs, and is finally divided by $\bar{T}^*$. Results are shown in the left-hand figure of Fig. 3. The same set of simulations is performed for a non-uniform spatial density; in particular we consider $\delta = 0.6$. Results are shown in the right-hand figure of Fig. 3.

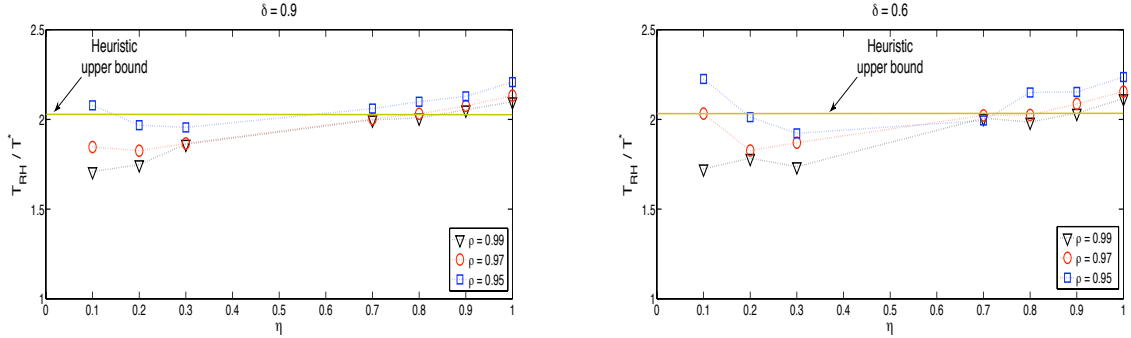

Fig. 3. Left Figure: Ratio between experimental system time under the RH policy and $\bar{T}^*$ in the case of uniform density (i.e., $\delta = 0.9$). Right Figure: Ratio between experimental system time under the RH policy and $\bar{T}^*$ in the case of non-uniform density ($\delta = 0.6$).

The results confirm that the system time $\bar{T}_{RH}$ follows the $\lambda|\mathcal{Q}|/(1-\varrho)^2$ growth predicted by Theorem 5.1. In Section V we argued that in heavy load $T_{RH}(\eta) \leq 2\bar{T}^*$ (see Eq. (18)) for all $\eta \in (0, 1)$; from Fig. 3, it can be observed that as $\varrho$ approaches one this asymptotic bound seems to be confirmed, both for a uniform $f$ and a non-uniform $f$. Finally, we also argued in Section V that the performance of the RH policy should improve as the horizon becomes shorter: this is indeed the case, in particular simulation results show that optimal performance is achieved for $\eta \approx 0.2$.

### C. Comparison between DC policy and RH policy

The RH policy should be compared with the version of the DC policy that is also adaptive with respect to $f$, i.e., with the DC policy with $r = 1$. Indeed, as observed before, the DC policy with $r = 1$ is the same as the RH policy when we set $\eta = 1$. Therefore, we can compare the two policies by looking at Fig. 3. One should note that the RH policy with $\eta \approx 0.2$ performs consistently better than the DC policy with $r = 1$, in particular it decreases the system time by a factor $\approx 1.25$.

### D. Execution of the Multi-Vehicle DC Policy

The $m$-DC policy adopts a partitioning scheme to distribute the workload among the agents; we refer the interested reader to [34] for a discussion about the performance of the distributed partitioning algorithm that the $m$-DC policy uses. Here, we only show a snapshot of a simulation experiment with 8 vehicles; the vehicles operate in the unit square with a uniform density $f$, and execute the $m$-DC policy with $r = 1$. The load factor is $\varrho = 0.8$;

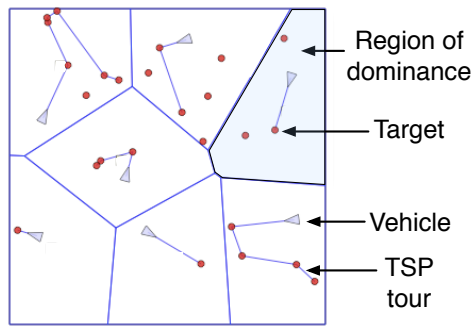

Fig. 4.  Example of execution of the $m$-DC policy, $m = 8$ , $\varrho = 0.8$, $\delta = 0.9$.

TABLE I

Unbiased Policies for the 1-DTRP

| Properties | DC Policy, $r \to \infty$ | DC Policy, $r = 1$ | RH Policy | UTSP Policy, $r \to \infty$ |
|---|---|---|---|---|
| Light-load performance | optimal | optimal | optimal | not optimal |
| Heavy-load performance | optimal | within 100 % of optimum | within 100% of optimum* | optimal |
| Adaptive to $\lambda$, $\bar{s}$, and $v$ | yes | yes | yes | no |
| Adaptive to $f$ | no | yes | yes | no |

TABLE II

Unbiased Policies for the $m$-DTRP

| Properties | DC Policy, $r \to \infty$ | UTSP Policy, $r \to \infty$ |
|---|---|---|
| Light-load performance | "locally" optimal | not optimal |
| Heavy-load performance | optimal for uniform $f$, within $m$ of optimal in general | optimal |
| Adaptive to $\lambda$, $\bar{s}$, and $v$ | yes | no |
| Adaptive to $f$ | no | no |
| Distributed | yes | no |

## VIII. Conclusion

The DTRP is a widely applicable model for dynamic task-assignment problems in robotic newtorks. In this paper, our objective was to design unbiased policies for the DTRP that (i) are adaptive (in particular do *not* require the knowledge of the arrival rate $\lambda$ and the statistics of the on-site service time), (ii) enjoy provable performance guarantees (in particular, are provably stable under any load condition), and (iii) are distributed. To this end, we introduced two new policies for the 1-DTRP, namely the DC policy and the RH policy, that were subsequently extended to the multi-vehicle case through the idea of partitioning policies.

Tables I and II provide a synoptic view of the results we achieved; in particular, our policies are compared with the best unbiased policy available in the literature, i.e., the UTSP policy with $r \to \infty$. In Table I, an asterisk * signals that the result is heuristic.

This paper leaves numerous important extensions open for further research. First, although unbiased service seems to be a natural constraint in most applications, in some cases this constraint might be relaxed and it is then of interest to design optimal or near-optimal *biased* policies that are adaptive and distributed (recall that allowing biased service may result in a strict reduction of the optimal mean system time for any non-uniform density $f$). Second, finding an unbiased policy for the 1-DTRP that is optimal in heavy load and does not rely on the knowledge of $f$ is both practically important and theoretically interesting. Third, the $m$-DC policy requires the knowledge of $f$ for the partition of the environment among the vehicles: we plan to design asynchronous distributed algorithms for equitable environment partitioning that do not rely on the knowledge of $f$. Finally, in our analysis, we considered omni-directional vehicles with first order dynamics: non-holonomic constraints will have to be taken into account for practical application to UAVs or other systems.

## Acknowledgments

The authors thank Stephen L. Smith for insightful feedback during the development of this work. The research leading to this work was partially supported by the Air Force Office of Scientific Research through grant AFOSR MURI-FA9550-07-1-0528. Any opinions, findings, and conclusions or recommendations expressed in this publication are those of the authors and do not necessarily reflect the views of the supporting organizations.